\pdfoutput=1
\documentclass[10pt,twocolumn,letterpaper]{article}

\usepackage[pagenumbers]{cvpr}

\usepackage{cuted} 
\usepackage{enumitem}

\usepackage{currfile} 

\usepackage{caption}

\renewcommand{\paragraph}[1]{\vspace{.45em}\noindent\textbf{#1}\enspace}

\usepackage{multirow}
\usepackage{booktabs}
\usepackage{amssymb}
\usepackage{graphicx}
\usepackage{bm}
\usepackage{xspace}

\DeclareMathOperator{\softmax}{softmax}
\newcommand{\cx}{c_x}
\newcommand{\cy}{c_y}

\newcommand{\dataset}{Kiss3000\xspace}

\newcommand{\baseLayers}{32\xspace} 
\newcommand{\baseVaeFactor}{16\xspace}

\newcommand{\baseMaxRefs}{10\xspace}

\newcommand{\nMannImg}{915\xspace}           
\newcommand{\nMannStyledImg}{600\xspace}     
\newcommand{\nMannLoose}{315\xspace}         
\newcommand{\nMannStyles}{84\xspace}         
\newcommand{\mannRes}{$3500{\times}5250$\xspace} 
\newcommand{\mannMP}{34.7\xspace}            
\newcommand{\nModelImg}{2274\xspace}         
\newcommand{\nModelStyles}{375\xspace}       
\newcommand{\modelRes}{$1500{\times}2000$\xspace} 
\newcommand{\modelMP}{3.0\xspace}            
\newcommand{\nModelDetail}{144\xspace}       
\newcommand{\nMannMacro}{215\xspace}         
\newcommand{\nPairCand}{267\xspace}          
\newcommand{\nPairs}{38\xspace}              
\newcommand{\nPairStyles}{21\xspace}         
\newcommand{\nPairsSameColor}{10\xspace}     
\newcommand{\nPairsSameColorStyles}{7\xspace} 
\newcommand{\nPairsCode}{14\xspace}          
\newcommand{\nPairsVLMOnly}{24\xspace}       
\newcommand{\nNegCtrl}{10\xspace}            
\newcommand{\nNegRejected}{10\xspace}        
\newcommand{\nTrainStyles}{43\xspace}        
\newcommand{\nValStyles}{20\xspace}          
\newcommand{\nTestStyles}{21\xspace}         
\newcommand{\nTestUsed}{20\xspace}           
\newcommand{\nModelOnlyStyles}{339\xspace}   
\newcommand{\nModelOnlyImg}{2000\xspace}     

\newcommand{\sMin}{\ensuremath{1/16}\xspace}        
\newcommand{\macroTag}{\ensuremath{1/16}\xspace}    
\newcommand{\detailTag}{\ensuremath{1/4}\xspace}    
\newcommand{\pModel}{\ensuremath{0.5}\xspace}       
\newcommand{\nFourier}{8\xspace}         
\newcommand{\lambdaInit}{\ensuremath{0.1}\xspace}   
\newcommand{\hfAlpha}{\ensuremath{2.0}\xspace}      
\newcommand{\gradQuantile}{99th\xspace}

\newcommand{\trainRes}{\ensuremath{768{\times}1024}\xspace} 
\newcommand{\loraRank}{16\xspace}        
\newcommand{\refRes}{768\xspace}         
\newcommand{\nMacroRefs}{2\xspace}

\newcommand{\sFull}{\ensuremath{1}\xspace}          
\newcommand{\sHalf}{\ensuremath{0.5}\xspace}        
\newcommand{\sLocal}{\ensuremath{0.125}\xspace}     
\newcommand{\scaleJitter}{15\%\xspace}   %
\newcommand{\nCropsFull}{4\xspace}       
\newcommand{\nCropsHalf}{12\xspace}      
\newcommand{\nCropsLocal}{24\xspace}     
\newcommand{\coverLocal}{70\%\xspace}    
\newcommand{\coverHalf}{50\%\xspace}     
\newcommand{\cropSide}{448\xspace}       
\newcommand{\nSelfTest}{six\xspace}      
\newcommand{\capSide}{1536\xspace}       
\newcommand{\blurA}{2\xspace}            
\newcommand{\blurB}{4\xspace}            
\newcommand{\poseTol}{15\%\xspace}       
\newcommand{\poseRot}{10\xspace}         
\newcommand{\poseScale}{0.8\xspace}

\definecolor{cvprblue}{rgb}{0.21,0.49,0.74}
\usepackage[pagebackref,breaklinks,colorlinks,allcolors=cvprblue]{hyperref}

\def\confName{CVPR}
\def\confYear{2027}

\title{Scale-Aware Reference Attention for\\Mannequin-to-Model Gown Generation}

\author{Guang Yang$^{1}$ \qquad Fengchen Liu$^{2}$\\[0.4em]
$^{1}$University of California, Los Angeles \qquad $^{2}$University of California, Berkeley}

\begin{document}
\maketitle
\begin{figure*}[t]
\centering
\includegraphics[width=\linewidth]{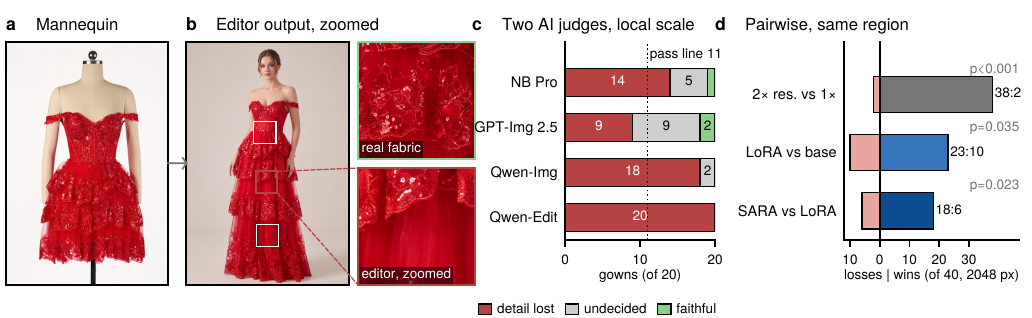}
\caption{Overview.
(a) Mannequin photo of a test gown.
(b) Output of Nano Banana Pro at 4K; boxes mark the local-scale crops seen by the judges, and one is enlarged next to the real fabric of the same gown.
(c) Labels on which both VLM judges agree, per editor over 20 test gowns; the dotted line is the pre-registered threshold.
(d) Pairwise judgments on the same garment region at $1536{\times}2048$: base model at $2\times$ resolution against itself, LoRA fine-tuning against the base model, and SARA against LoRA fine-tuning (wins:losses of 40 items).}
\label{fig:teaser}
\end{figure*}

\begin{abstract}
Formal gowns are sold online with photos of models wearing them, and buyers zoom in to inspect sequins, lace and beading.
Image editors can turn a photo of a gown on a mannequin into a convincing photo of a person wearing it, but they receive every reference photo without any notion of its magnification, and the fine detail of the gown is lost when the output is viewed up close.
We propose scale-aware reference attention (SARA), which tells an open diffusion transformer the zoom level of the output and of every reference photo, biases the attention of the output toward references of a similar zoom level, uses macro photos of the fabric as references, and is trained on a mix of whole-garment and zoomed targets.
To train and evaluate it, we introduce \dataset, a dataset of 915 high-resolution mannequin photos and 2274 on-model photos of formal gowns with cross-domain pairs and splits by style, and zoom-crop evaluation, which compares garment crops at full, half and local scale without pixel alignment and judges local fidelity with two vision-language models gated on control crops.
With this protocol, Nano Banana Pro at 4K, the highest-resolution commercial editor we test, loses or redraws local detail on 14 of 20 test gowns.
At $1536{\times}2048$ pixels, both judges prefer SARA over a LoRA fine-tuning baseline on 18 test items and the baseline on 6 ($p=0.023$), and over the base model on 26 against 4 ($p=5.9{\times}10^{-5}$), while SARA keeps the gown design as often as the baseline.
\end{abstract}

\section{Introduction}
\label{sec:intro}

Online stores present formal gowns, such as prom, bridesmaid and quinceañera dresses, with photos of models wearing them.
Much of the appeal of these garments lies in their decoration: sequins, lace, beading and embroidery.
Listings therefore pair overview photos with close-ups, and shoppers seek close-up views of fine garment detail~\cite{garmentzoom,detailanywhere}.
Photos with real models require substantial effort and cost, whereas mannequin photos are more affordable~\cite{m2hvideo}.
Turning mannequin photos into on-model photos is thus an attractive use of image generation, and it is close to virtual try-on~\cite{vitonhd,idmvton,catvton,fitdit} and to reference-conditioned image editing~\cite{fluxkontext,qwenimage}.
For a decorated gown, a convincing full view is not enough: the product is the fabric, and a buyer who zooms in should see the sequins and the lace of this gown, not a plausible substitute.

We observe that current reference-conditioned editors are blind to scale.
A gown is typically photographed as a full-body shot and as macro shots of the fabric, at very different magnifications, yet an editor places all references in one token sequence and tells them apart by a positional coordinate, with no notion of magnification~\cite{fluxkontext,qwenimage}.
When it draws a full-body view, it cannot tell which reference shows the fabric at the scale it is drawing, and under zoom the decoration is redrawn rather than transferred.

We address this with scale-aware reference attention (SARA, \cref{fig:teaser}).
SARA conditions the generator on the zoom level of the output, tags every reference with its own zoom level, and biases the attention from the output to each reference by the log-scale gap between them, so that macro references guide fine structure and full-body references guide the overall garment.
It adds one parameter per transformer block, reduces exactly to the base model at initialization, and keeps fused attention kernels usable.
We train SARA on a mix of whole-garment and zoomed targets with a loss weighted toward fine structure; the whole-garment targets keep the full view intact.

Training and evaluating such a model needs data and a measure that see fine detail.
A generated model photo is not pixel-aligned with the mannequin photos, so full-reference metrics such as SSIM and LPIPS~\cite{ssim,lpips} confound changes in pose and layout with changes in garment appearance, and try-on benchmarks report distribution metrics such as FID and KID~\cite{fid,kid} on whole images~\cite{vitonhd,dresscode}, which do not match garment regions across magnifications.

We make the following contributions.
\begin{itemize}[leftmargin=*,itemsep=1pt,topsep=2pt]
\item We propose SARA, a scale-aware reference attention for reference-conditioned diffusion transformers (\cref{sec:method}).
\item We introduce \dataset, a dataset of high-resolution mannequin photos (median 34.7 megapixels) and on-model photos of formal gowns, with cross-domain pairs and splits by style (\cref{sec:dataset}).
\item We propose zoom-crop evaluation, an alignment-free protocol that compares garment crops at full, half and local scale. A self-test with controlled degradations shows that common automatic measures, including SSIM and LPIPS, fail at local scale, so we judge local fidelity with two vision-language model (VLM) judges from different families, gated on control crops (\cref{sec:evaluation}).
\item With this protocol, Nano Banana Pro at 4K loses or redraws local detail on 14 of 20 test gowns (\cref{sec:e0}). At $1536{\times}2048$ pixels, both judges prefer SARA over a LoRA fine-tuning baseline~\cite{lora} on 18 test items and the baseline on 6, and over the base model on 26 against 4, while SARA keeps the gown design as often as the baseline (\cref{sec:study}).
\end{itemize}

\section{Related Work}
\label{sec:related}

\paragraph{Virtual try-on.}
Diffusion try-on methods condition a denoising U-Net on the garment through a parallel encoder or warped features~\cite{tryondiffusion,stableviton,ootdiffusion,idmvton}, concatenate person and garment in one input~\cite{catvton}, move to diffusion transformers~\cite{fitdit,any2anytryon}, or learn attention flow between the two images~\cite{leffa}.
FitDiT adds a frequency-domain loss for high-frequency garment detail~\cite{fitdit}, and WearWow generates at $2048{\times}1536$ and uses preference rewards against over-smoothed fabric~\cite{wearwow}.
These methods take whole garment images and are evaluated on full views, typically on VITON-HD and DressCode~\cite{vitonhd,dresscode}.

\paragraph{Garment-centric and mannequin-to-model generation.}
Other work generates a person wearing a given garment without a source person photo~\cite{magicclothing,imagdressing,anydressing}.
GCO and FashionMAC segment the garment from a dressed mannequin or person and outpaint a model around it~\cite{gco,fashionmac}, and M2HVideo turns mannequin footage into videos of a human model at $512{\times}256$ pixels~\cite{m2hvideo}.
None of these works conditions on the scale of a view or evaluates the garment at several magnifications.

\paragraph{Close-up and zoomable garment imagery.}
Eevee provides garment close-ups and close-up videos for video try-on; its baseline feeds the close-ups as extra reference frames, and its garment consistency metric pools DINOv2 features over the whole garment~\cite{eevee}.
ACDG-VTON regenerates a zoomed region by cropping and upsampling its pixel-aligned control image~\cite{acdgvton}.
DetailAnywhere generates a close-up of a region marked on a product photo~\cite{detailanywhere}.
UltraZoom and GarmentZoom super-resolve a full-view photo with unaligned close-ups of the same object, GarmentZoom over a continuous range of zoom factors with sliding-window inference~\cite{ultrazoom,garmentzoom}.
The Qwen-Image-2.0 report observes that generators often introduce repeated textures and lose fine-grained detail at 2K resolution and above~\cite{qwenimage2}, and on a try-on benchmark a general-purpose editor scores high on realism but lower on texture~\cite{openvtonbench}.
These works generate or refine detail from close-ups; SARA instead tells the generator at which magnification each reference and the output are, and we measure fidelity at matched magnifications.

\paragraph{Evaluating garment fidelity.}
Full-reference metrics such as SSIM and LPIPS compare corresponding spatial locations~\cite{ssim,lpips}, and FID and KID compare distributions of whole images~\cite{fid,kid}.
Texture statistics~\cite{gatys,dists}, position-free matching of deep features~\cite{contextualloss,dinov2}, and local feature matching~\cite{sift} tolerate misalignment.
For garments, fidelity has been measured with LPIPS on patches around matched points~\cite{magicclothing}, with DINO features of masked garment crops~\cite{omnitry} or of soft-masked, pooled garment features~\cite{eevee}, with FID and KID on randomly sampled patches~\cite{ultrazoom}, and with DINOv3 features over progressively eroded garment masks against a paired ground truth~\cite{openvtonbench}.
Recent benchmarks also score try-on results with vision-language judges~\cite{openvtonbench,vteditbench,dat}.
None of these protocols compares crops at several matched magnifications against unaligned references.
Our zoom-crop protocol does so, tests each measure against controlled degradations before use (\cref{sec:selftest}), and complements the measures with two vision-language judges gated on control items (\cref{sec:judged}).

\paragraph{Scale and reference conditioning.}
AnyResGAN and ScaleSpaceGAN condition a generator on patch scale and coordinates~\cite{anyresgan,scalespacegan}, Patch Diffusion on patch location and size~\cite{patchdiffusion}, and ZoomLDM on magnification~\cite{zoomldm}; JewelTry encodes the physical size of a product in a scale token~\cite{jeweltry}.
Other work generates consistent zoom sequences or exceeds the native resolution without training~\cite{powersoften,chainofzoom,multidiffusion,demofusion,scalecrafter}.
Reference-based super-resolution transfers texture from a high-resolution reference through correspondence matching~\cite{srntt,ttsr,c2matching,datsr} or diffusion~\cite{refir}.
Reference-conditioned editors place reference tokens in the same sequence as the target~\cite{ominicontrol,easycontrol,fluxkontext,qwenimage} and tell reference images apart by an extra positional coordinate, without an explicit magnification variable~\cite{fluxkontext,qwenimage}; OminiControl adds one user-set bias to the attention logits between target and condition tokens~\cite{ominicontrol}.
SARA instead biases the attention to each reference by its log-scale gap to the generated view, a distance bias in the spirit of ALiBi~\cite{alibi}.

\section{Scale-Aware Reference Attention}
\label{sec:method}

SARA makes the zoom level explicit for the output and for every reference of an open reference-conditioned diffusion transformer: it conditions the generator on the target view (\cref{sec:scalecond}), biases attention by the scale gap between output and reference (\cref{sec:sara}), and is trained with whole-garment and zoomed targets and a detail-weighted loss (\cref{sec:training}).

\begin{figure}[t]
\centering
\includegraphics[width=\linewidth]{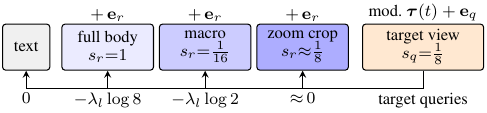}
\caption{Scale-aware reference attention in one block, for a target view at $s_q=1/8$.
Each reference carries a scale tag $s_r$ and a tag embedding $\mathbf{e}_r$; the target view enters the modulation of the target tokens.
The attention logits from target queries to the keys of reference $r$ receive the bias $-\lambda_l|\log s_q-\log s_r|$ (\cref{eq:bias}); text and target keys are not biased.
With $\lambda_l=0$ the block is the base block.}
\label{fig:method}
\end{figure}

\subsection{Base model, views and scales}
\label{sec:base}
\label{sec:views}
We build on Qwen-Image-2.1~\cite{qwenimage21}, an open diffusion transformer with \baseLayers single-stream blocks that conditions on up to \baseMaxRefs reference images.
Its autoencoder downsamples by \baseVaeFactor in each dimension, so one token covers a $\baseVaeFactor{\times}\baseVaeFactor$ pixel patch.
As in other reference-conditioned editors~\cite{fluxkontext,qwenimage}, text tokens, the tokens of each reference and the target tokens form one sequence; reference images are distinguished by an extra positional coordinate, and no input encodes their magnification.
Attention is block-causal: target tokens attend to the text and to all references, and reference tokens never attend to the target.
We fine-tune with low-rank adapters~\cite{lora} on the attention and feed-forward projections of every block and with the flow-matching objective~\cite{flowmatching,rectifiedflow}: for $\mathbf{x}_t=(1-t)\,\mathbf{x}_0+t\,\bm{\epsilon}$, the network $\mathbf{u}_\theta$ predicts the velocity $\bm{\epsilon}-\mathbf{x}_0$.

Let $B$ be the bounding box of the garment in an image and $H_B$ its height.
A view is a crop with the aspect ratio of the output, described by $v=(\cx,\cy,s)$: the center in coordinates normalized by the width and height of $B$, and the scale $s$, the crop height divided by $H_B$.
The full garment has $s=1$, and a $2\times$ zoom is a fixed step in $\log s$.
The input is a set of mannequin photos $\mathcal{R}=\{(I_r,s_r)\}_{r=1}^{R}$ with scale tags $s_r$.
A full-body photo has $s_r=H_I/H_B\ge 1$, where $H_I$ is the image height.
A macro photo shows no garment box and receives the nominal tag $s_r=\macroTag$.
The output is an on-model photo in a target view $v_q=(\cx,\cy,s_q)$.

\subsection{Scale conditioning and scale tags}
\label{sec:scalecond}
The generator is conditioned on the target view through
\begin{equation}
  \mathbf{e}_q=\mathrm{MLP}_v\big(\left[\gamma(\cx),\,\gamma(\cy),\,\gamma(\log s_q)\right]\big),
  \label{eq:viewemb}
\end{equation}
where $\gamma$ maps a scalar to sine and cosine features at \nFourier frequencies together with the scalar itself, and $\mathrm{MLP}_v$ has two layers.
The base model computes one modulation vector from the timestep embedding $\bm{\tau}(t)$.
We add $\mathbf{e}_q$ to the conditioning of the target tokens only, which are then modulated by $\bm{\tau}(t)+\mathbf{e}_q$; text and reference tokens keep their modulation, so their cache stays valid.
Each reference also receives an embedding of its own tag, $\mathbf{e}_r=\mathrm{MLP}_s(\gamma(\log s_r))$, added to all of its tokens at the input of the transformer.
The last layers of both MLPs are initialized to zero, so training starts from the base model.

\subsection{Scale-aware reference attention}
\label{sec:sara}
Scale conditioning says which view to produce, but not which reference shows that view at a comparable magnification.
Let $\mathcal{K}_r$ be the key positions of reference $r$.
In block $l$, the attention of a target query $i$ to a key $j$ is
\begin{equation}
  a_{ij}=\softmax_j\!\left(\frac{\mathbf{q}_i^{\top}\mathbf{k}_j}{\sqrt{d}}+b^{l}_{ij}\right),
  \label{eq:attn}
\end{equation}
\begin{equation}
  b^{l}_{ij}=
  \begin{cases}
    -\lambda_l\,\big|\log s_q-\log s_r\big| & \text{if } j\in\mathcal{K}_r,\\
    0 & \text{otherwise},
  \end{cases}
  \label{eq:bias}
\end{equation}
where $d$ is the head dimension and $\lambda_l\ge 0$ is one learnable strength per block, shared by its heads.
Only rows of target queries are biased; text and target keys receive no bias.
Because the bias is constant over the keys of one reference, it leaves the distribution of attention inside a reference unchanged and scales the total attention paid to reference $r$ by $\exp(-\lambda_l|\log s_q-\log s_r|)$ before normalization.
A macro reference therefore receives relatively more attention when a macro view is generated and less when the full body is generated (\cref{fig:method}).

Three properties follow.
First, with $\lambda_l=0$ the block computes the base attention, and with zero-initialized embeddings and adapters the whole model equals the base model; we initialize $\lambda_l$ at \lambdaInit and project it onto $\lambda_l\ge 0$ after every optimizer step.
Second, the bias depends only on scalars known before sampling and adds one parameter per block; it is applied exactly through one extra query and key channel, so fused attention kernels remain usable.
Third, reference tokens never attend to the target, so their keys and values stay cached.

\subsection{Training}
\label{sec:training}
\paragraph{Training samples.}
Gowns with photos in both domains are reserved for testing (\cref{sec:dataset}), so no training gown provides a mannequin input together with an on-model target.
We therefore train on two kinds of samples, each drawn with probability \pModel.
(a) Mannequin view reconstruction: the target is a view of one non-macro mannequin photo of a training gown, and the references are another non-macro mannequin photo of the same gown and its macro photos.
(b) On-model reconstruction: the target is a view of one on-model photo, and the references are another on-model photo of the same gown and its close-ups, tagged $s_r=\detailTag$.
On-model photos have no reliable garment mask, so their garment box is the whole image.
With probability $0.5$ the target is the whole garment ($s=1$); otherwise its scale is drawn with $\log s$ uniform in $[\log s_{\min},0]$, where $s_{\min}$ is the larger of \sMin and the smallest scale at which the crop is not upsampled, and its center is a random garment pixel.
The whole-garment targets matter: with zoomed targets only, the model learns to return a zoomed-in crop when asked for the full view (\cref{sec:ablation}).
A crop of the reference photo around the same garment coordinates, at a scale jittered around $s_q$, is added as a further reference with its own tag unless that scale exceeds $0.5$.
A fixed instruction names the kind of output and never mentions the scale, so the tags are the only scale signal.
Noise levels follow the resolution-dependent shift of the base sampler.

\paragraph{Detail-weighted flow matching.}
Most of a gown is plain fabric, so decoration contributes little to a uniform loss.
We weight each latent position $p$ by
\begin{equation}
  w_p = 1+\alpha\, \hat{g}_p\, m_p,
  \label{eq:weight}
\end{equation}
where $\hat{g}_p$ is the Sobel gradient magnitude of the target luminance averaged over the patch of $p$, divided by its \gradQuantile percentile inside the garment and clipped at one, and $m_p$ is the garment fraction of the patch ($m_p=1$ for on-model targets).
With $\bar{w}_p=w_p/\sum_{p'}w_{p'}$ and $C$ latent channels, the loss is
\begin{equation}
  \mathcal{L}=\mathbb{E}_{t,\mathbf{x}_0,\bm{\epsilon}}\Big[\frac{1}{C}\sum_p \bar{w}_p\,\big\|\mathbf{u}_{\theta,p}-(\bm{\epsilon}-\mathbf{x}_0)_p\big\|_2^2\Big],
  \label{eq:loss}
\end{equation}
where $\mathbf{u}_{\theta,p}$ is predicted from $(\mathbf{x}_t,t,\mathcal{R},v_q)$; we use $\alpha=\hfAlpha$, and $\alpha=0$ gives the standard objective.

\paragraph{Configuration.}
SARA uses up to \nMacroRefs macro or close-up references and one zoom-crop reference, adapters of rank \loraRank on the attention and feed-forward projections of every block, outputs of \trainRes pixels (width by height) and references resized to an area of $\refRes^2$ pixels; it is trained for 4000 steps.
At test time we generate at $1536{\times}2048$ pixels.

\section{The \dataset Dataset}
\label{sec:dataset}

\paragraph{Source and content.}
\dataset is a collection of product photos of formal gowns, donated for research by the company that owns them.
It has two domains (\cref{tab:dataset}).
The mannequin domain has \nMannImg photos, \nMannStyledImg of which belong to \nMannStyles identified styles; the other \nMannLoose are not assigned to a style and are excluded from all splits.
Mannequin photos are large, with a median of \mannMP megapixels, and a vision-language model labels each of them as full body (375), back (271), macro (215) or half body (54).
An AI agent, not a human, spot-checked 60 random labels and agreed with 58 (2 borderline).
The on-model domain has \nModelImg photos of \nModelStyles styles, most of them \modelRes pixels (\modelMP megapixels), grouped by occasion into homecoming, prom, bridesmaid and quinceañera gowns; \nModelDetail of them are close-ups.
A mannequin photo thus has more than ten times the pixels of an on-model photo. The two domains are photographed independently, so pose, viewpoint and framing differ, and one design is often shown in different colors.

\begin{table}[t]
\centering
\footnotesize
\caption{Statistics of \dataset. Pairs link a mannequin style to an on-model style of the same design; splits are by mannequin style.}
\label{tab:dataset}
\setlength{\tabcolsep}{4pt}
\begin{tabular}{@{}lcc@{}}
\toprule
 & Mannequin & On-model \\
\midrule
Photos (unassigned) & \nMannImg{} (\nMannLoose) & \nModelImg \\
Styles & \nMannStyles & \nModelStyles \\
Median size (MP) & \mannMP & \modelMP \\
Most common size (px) & \mannRes & \modelRes \\
Macro / close-up photos & \nMannMacro & \nModelDetail \\
\midrule
Design-level pairs (styles) & \multicolumn{2}{c}{\nPairs{} (\nPairStyles)} \\
\quad same color (styles) & \multicolumn{2}{c}{\nPairsSameColor{} (\nPairsSameColorStyles)} \\
\midrule
Train / val / test styles & \multicolumn{2}{c}{\nTrainStyles{} / \nValStyles{} / \nTestStyles} \\
On-model-only styles (train) & \multicolumn{2}{c}{\nModelOnlyStyles{} (\nModelOnlyImg{} photos)} \\
\bottomrule
\end{tabular}
\end{table}

\paragraph{Cross-domain pairs.}
A pair links a mannequin style to an on-model style of the same design.
Rules on the style identifiers propose \nPairCand candidates, and a visual retrieval sweep, in which a vision-language model looks for each on-model style among contact sheets of all mannequin styles, proposes further candidates; the sweep recovers 12 of 14 known pairs.
A vision-language model then checks every candidate on a contact sheet that shows no identifier, in two independent passes with different photos, and a pair is accepted only if both passes agree.
The check rejected \nNegRejected of \nNegCtrl mismatched negative controls.
The result is \nPairs design-level pairs covering \nPairStyles mannequin styles, of which \nPairsSameColor pairs (\nPairsSameColorStyles styles) also match in color.
For \nPairsCode pairs the identifiers agree as well, while \nPairsVLMOnly rest on the model check alone and have not been verified by humans.
Our evaluations use only mannequin photos of the test gowns as inputs and references, so they do not depend on the accuracy of the pairs.

\paragraph{Splits and identifiers.}
We split by style, never by image.
The test split contains every mannequin style with a pair, \nTestStyles styles; \nTestUsed of them have a full-body mannequin photo and form the test set of \cref{sec:experiments}.
The validation split has \nValStyles randomly drawn mannequin styles and the training split the remaining \nTrainStyles.
A further \nModelOnlyStyles on-model styles (\nModelOnlyImg photos) are not linked to any mannequin style and are used only for training.
Gowns are referred to by anonymous identifiers (G01, G02, \dots), and all people shown in our figures are generated.

\section{Zoom-Crop Evaluation}
\label{sec:evaluation}

Since generated and mannequin photos are not aligned (\cref{sec:intro}), we compare garment crops at matched scales, first with automatic measures that need no alignment and, because these turn out to be unreliable at local scale, with judged comparisons under controls.

\subsection{Crop sampling and measures}
\label{sec:crops}
\label{sec:measures}
We segment the garment with a SegFormer human parser fine-tuned on ATR~\cite{segformer,atr} and fall back to GrabCut~\cite{grabcut} when the parser finds almost no garment.
Let $B$ be the longer side of the garment box.
We sample square crops of side $sB$ at three scales, full ($s=\sFull$), half ($s=\sHalf$) and local ($s=\sLocal$), with a log-uniform jitter of $\pm$\scaleJitter.
Half and local crops are centered on random garment pixels and kept only if at least \coverHalf and \coverLocal of the crop, respectively, is garment; full crops are centered on the garment box.
We draw \nCropsFull, \nCropsHalf and \nCropsLocal crops per image at the three scales, from the output and from every reference photo (the other non-macro mannequin photos of the gown, never an input), resize them to $\cropSide{\times}\cropSide$ pixels and ignore pixels outside the garment mask.
Four measures compare the crops of an output with the reference crops of the same scale.
Dense feature similarity (DFS) matches every DINOv2 patch feature~\cite{dinov2} of an output crop to its most similar reference patch and averages the cosine similarities.
We define texture statistics distance (TSD) as the normalized distance between masked Gram matrices~\cite{gatys} of four VGG-16 layers~\cite{vgg} to the closest reference crop.
Local feature matching (LFM) is the inlier ratio of a homography fit to ratio-tested SIFT matches~\cite{sift} with the best-matching reference crop.
The high-frequency ratio (HFR) divides the Laplacian energy of an output crop by the median energy of the reference crops; it detects blur but not wrong structure.

\begin{table}[t]
\centering
\footnotesize
\caption{Self-test of detail measures at local scale (mean over the six self-test gowns). The original is a real mannequin photo, compared with a resolution cap, two blurs, a different gown and a pose change (relative change). Arrows give the better direction. A measure passes if it satisfies the five pre-registered checks C1 to C5 (\cref{sec:selftest}). LPIPS and SSIM are computed on whole images.}
\label{tab:selftest}
\setlength{\tabcolsep}{2.2pt}
\begin{tabular}{@{}lcccccrl@{}}
\toprule
Measure & Orig. & Cap & Blur 2 & Blur 4 & Other & Pose & Fails \\
\midrule
DFS $\uparrow$  & 0.821 & 0.784 & 0.806 & 0.773 & 0.614 & $-0.5\%$  & C2 \\
TSD $\downarrow$ & 0.295 & 0.706 & 0.687 & 0.916 & 0.956 & $+8.8\%$  & C2 \\
LFM $\uparrow$  & 0.324 & 0.202 & 0.171 & 0.010 & 0.080 & $-2.3\%$  & none \\
HFR $\uparrow$  & 1.162 & 0.026 & 0.024 & 0.003 & 1.990 & $-15.4\%$ & C4, C5 \\
\midrule
LPIPS $\downarrow$ & 0.236 & 0.299 & 0.291 & 0.312 & 0.328 & $+14.7\%$ & C2 \\
SSIM $\uparrow$  & 0.733 & 0.757 & 0.759 & 0.766 & 0.702 & $+2.1\%$  & C1, C2, C3 \\
\bottomrule
\end{tabular}
\end{table}

\subsection{Do the measures see detail loss?}
\label{sec:selftest}
A detail measure must rank degraded images below faithful ones and tolerate a change of pose; we tested this with checks fixed in advance.
For \nSelfTest decorated gowns, a three-quarter mannequin photo plays the role of a perfect output, and the other non-macro photos of the same gown are the references.
Degraded versions are a resolution cap (down to \capSide pixels on the long side and back), Gaussian blur with $\sigma=\blurA$ and $\sigma=\blurB$ pixels, the photo of a different gown, and a pose change (perspective warp, rotation by \poseRot degrees, rescaling by \poseScale).
At local scale, a measure must rank the original above the cap (C1), the cap at least as high as the weaker blur (C2), the weaker blur above the stronger one (C3), and the original above the different gown (C4), and it may change by at most \poseTol under the pose change (C5).

\cref{tab:selftest} shows the result.
Only LFM passes all five checks, and its C2 check holds in the mean but in only 2 of the 6 gowns.
DFS and TSD fail C2, and HFR fails C4 and C5: a different decorated gown has more high-frequency energy than the original.
SSIM and LPIPS, computed on whole images and tested under the same checks, also fail: SSIM ranks the capped and both blurred versions above the original, the strongest blur highest, and LPIPS fails C2.
For DFS, the degradations are invisible at full scale (0.890 for the original, 0.889 for the strongest blur) and separate only at local scale (0.821 against 0.773).
On generated outputs, LFM sits at its floor: at $1536{\times}2048$ pixels, its median at local scale over the test gowns is 0 for both the base model and the LoRA baseline.
Fidelity under zoom must thus be measured at local scale, where the automatic measures we tested cannot rank methods.

\subsection{Judged evaluation with controls}
\label{sec:judged}
We therefore judge local crops with two vision-language models (VLMs) from different families, Claude Opus 5.5 and GPT-6.1 (medium reasoning effort, default sampling, one call per panel); these are model judgments, not human ones.
Each panel shows the same-color mannequin macro photo of the gown as the reference and three local crops of a candidate, near three fixed points on the vertical center line of the garment box (at 20\%, 50\% and 80\% of its height), together with a thumbnail of the candidate that marks the crops.
Panels show no method, model or gown name, and both judges receive the same instruction.

\paragraph{Absolute labels.}
A judge labels a candidate as ``structure clearly wrong or blurred'', ``essentially faithful'' or ``cannot judge''.
The primary label of an item is the common label of the two judges, and ``cannot judge'' when they disagree; per method and gown, the labels of the single-reference and multi-reference inputs are merged in favor of the method: a gown counts as faithful if either input gives a faithful label, as wrong only if both are wrong, and as undecided otherwise.
A judge is used only if it labels control panels correctly in at least 18 of 20 cases per class: crops of another real photo of the same gown (faithful), the same crops blurred with $\sigma=4$ pixels (wrong), and crops of a different gown (wrong).

\paragraph{Pairwise comparisons.}
A panel shows two candidates X and Y in random order, each with three crops centered on the same three points with side $0.125$ times the garment-box height, and the full-body input photo below the reference.
The judge states which candidate's crops are closer to the reference (X, Y or tie) and, for each candidate, whether its thumbnail shows the same dress design.
An item counts as a win or a loss only when both judges agree, and we apply a two-sided exact sign test to wins against losses; the control is a real photo against a blurred copy of itself ($\sigma=4$), which each judge must resolve in at least 18 of 20 cases.
Because the crop side is tied to the garment box, a model that frames the garment larger gives the judge more source pixels per crop, so we compare models at the same output size and check that their framing is similar.

\section{Experiments}
\label{sec:experiments}

\begin{table}[t]
\centering
\footnotesize
\setlength{\tabcolsep}{2.4pt}
\caption{Editors under zoom: number of the 20 test gowns whose local-scale crops are judged clearly wrong or blurred (W), largely faithful (F) or undecided (U), with labels on which both judges agree, merged over the inputs with and without macro photos (\cref{sec:judged}).
The pre-specified rule is W\,$\ge$\,11.
Judge agreement on control panels: 60 of 60.}
\label{tab:e0}
\begin{tabular}{@{}lcc@{}}
\toprule
Editor & Output size & W / F / U \\
\midrule
Nano Banana Pro (4K) & $3392{\times}5056$ & \textbf{14} / 1 / 5 \\
GPT Image 2.5 (high) & $2480{\times}3312$ & \phantom{0}9 / 2 / 9 \\
Qwen-Image-2.1 & $1376{\times}2048$ & 18 / 0 / 2 \\
Qwen-Image-Edit-2511 & $1024{\times}1536$ & 20 / 0 / 0 \\
\bottomrule
\end{tabular}
\end{table}

\begin{table}[t]
\centering
\footnotesize
\setlength{\tabcolsep}{2.6pt}
\caption{Pairwise judgments of local-scale crops at $1536{\times}2048$ (last row: base model at both sizes) on 40 test items (20 gowns, 2 seeds), same garment region and crop size.
W/L: items on which both judges prefer the first or the second model; Dis.: the judges disagree; $p$: two-sided exact sign test.
Design: items both judges call the same design as the reference, first / second model (of 40).}
\label{tab:method}
\begin{tabular}{@{}lcccc c@{}}
\toprule
First vs.\ second model & W & L & Dis. & $p$ & Design \\
\midrule
\textbf{SARA} vs.\ LoRA & \textbf{18} & 6 & 16 & 0.023 & 38 / 38 \\
\textbf{SARA} vs.\ base & \textbf{26} & 4 & 10 & $5.9{\times}10^{-5}$ & 36 / 37 \\
LoRA vs.\ base & 23 & 10 & 7 & 0.035 & 38 / 35 \\
\midrule
Base, $2048$ vs.\ $1024$ & 38 & 2 & 0 & $1.5{\times}10^{-9}$ & 37 / 38 \\
\bottomrule
\end{tabular}
\end{table}

\subsection{Setup}
\label{sec:setup}
All experiments use the 20 test gowns that have a frontal full-body mannequin photo (\cref{sec:dataset}).
\paragraph{Commercial and open editors.}
We run Nano Banana Pro~\cite{nanobananapro} at its 4K setting and GPT Image 2.5~\cite{gptimage25} at high quality through their official interfaces, and the open Qwen-Image-2.1~\cite{qwenimage21} and Qwen-Image-Edit-2511 zero-shot with seed 0 (output sizes in \cref{tab:e0}).
Every editor receives the same instruction with the full-body photo alone or together with one or two macro photos of the same colorway.
\paragraph{SARA and baselines.}
SARA and the LoRA baseline are trained from Qwen-Image-2.1 on the training split with seed 0 and a checkpoint every 250 steps.
The LoRA baseline trains the same adapters with whole-garment targets, one full-body reference and the standard flow-matching loss, without view conditioning, scale tags, macro references or attention bias.
A checkpoint rule, fixed before any validation output was generated, uses only validation gowns: a checkpoint is eligible if the garment box of its full views stays clear of the bottom image edge on at least as many gowns as for the base model minus two, and the eligible checkpoint with the highest full-scale feature similarity is selected; it selects step 4000 for both models.
At test time, every model receives the frontal full-body photo (SARA also one macro photo), is asked for the full view, $s_q=1$, and generates two seeds per gown at $1536{\times}2048$ pixels.
\paragraph{Judging.}
Editors are labeled with absolute judgments and models are compared pairwise (\cref{sec:judged}); both judges label all 60 control panels of the editor study and all 20 control pairs of the pairwise study correctly.

\subsection{Commercial editors lose detail under zoom}
\label{sec:e0}
The rule for this study was fixed before any output was generated: detail loss is confirmed if Nano Banana Pro or GPT Image 2.5 is judged clearly wrong or blurred on at least 11 of the 20 gowns.
With the labels on which both judges agree, Nano Banana Pro at 4K is judged clearly wrong or blurred on 14 of 20 gowns and largely faithful on one (\cref{tab:e0}), and the outcome is the same when each judge is used alone and when only the full-body input or only the input with macro photos is considered.
Macro photos in the input do not remove the failure: Nano Banana Pro is judged wrong on 18 gowns with the full-body photo alone and on 15 with macro photos added.
\cref{fig:teaser}b shows an example: the full view is convincing, but under zoom the decoration differs from the fabric of the gown.
For GPT Image 2.5, the two judges disagree on many gowns and the agreed labels leave 9 of 20 undecided.
The open editors, which generate at lower resolution, are judged wrong on 18 and 20 gowns and faithful on none.
Current editors, including the strongest commercial one, therefore do not preserve fine gown detail at local scale.

\subsection{SARA against LoRA fine-tuning}
\label{sec:study}
\label{sec:main_result}
\begin{figure}[t]
\centering
\includegraphics[width=\linewidth]{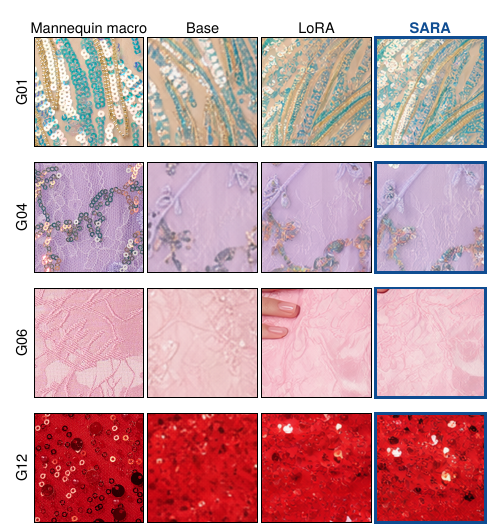}
\caption{The same garment region at local scale, as shown to the judges ($1536{\times}2048$ outputs, seed 0), next to the mannequin macro photo of each gown.
Rows are the first four test gowns, by id, on which both judges prefer SARA over the LoRA baseline.
SARA follows the motif and the density of the decoration more closely, for example the sequin stripes of G01 and the sequin scale of G12.}
\label{fig:zoomcrops}
\end{figure}

\begin{figure*}[t]
\centering
\setlength{\tabcolsep}{1pt}
\renewcommand{\arraystretch}{0.6}
\newcommand{\outimg}[1]{\includegraphics[height=0.95in]{fig/img/outputs/#1.jpg}}
\begin{tabular}{ccc@{\hspace{6pt}}cccc}
 & \multicolumn{2}{c}{\footnotesize without whole-garment targets} & & \multicolumn{3}{c}{\footnotesize $1536{\times}2048$} \\
 & \footnotesize scale cond.\ only & \footnotesize SARA & & \footnotesize Base & \footnotesize LoRA & \footnotesize \textbf{SARA} \\
\raisebox{0.42in}{\rotatebox{90}{\footnotesize G02}} & \outimg{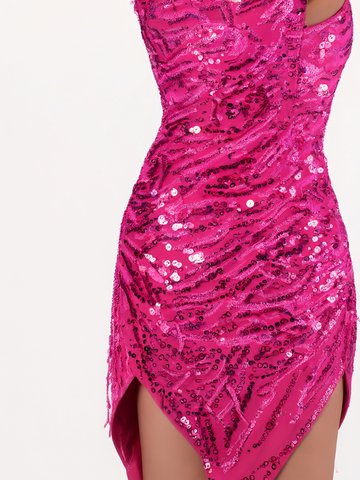} & \outimg{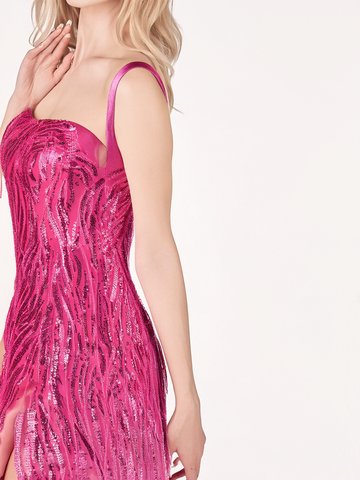} & & \outimg{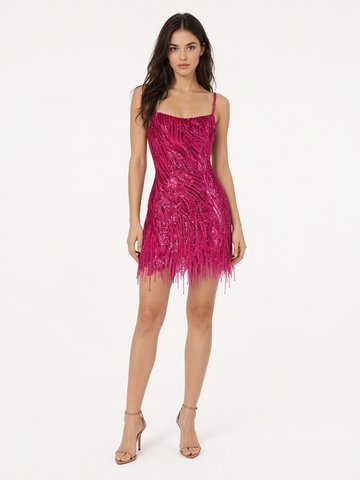} & \outimg{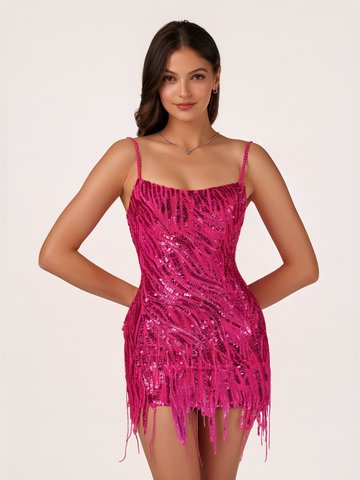} & \outimg{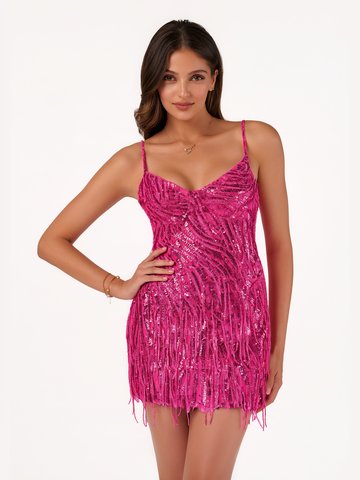} \\
\raisebox{0.42in}{\rotatebox{90}{\footnotesize G05}} & \outimg{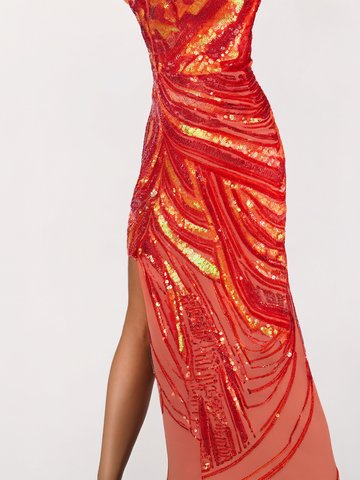} & \outimg{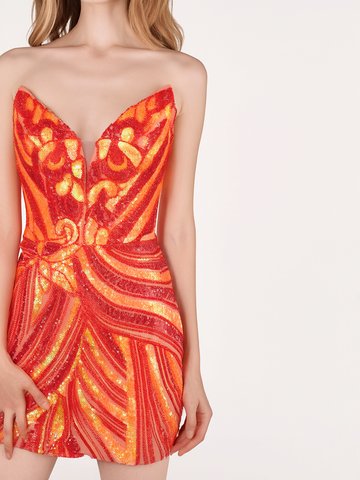} & & \outimg{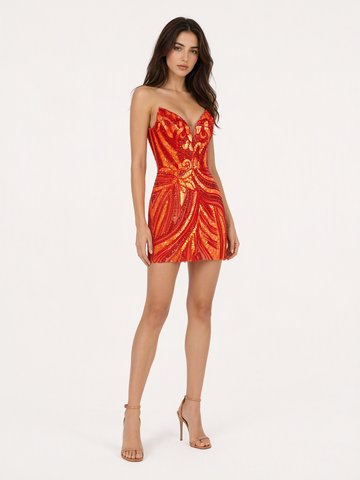} & \outimg{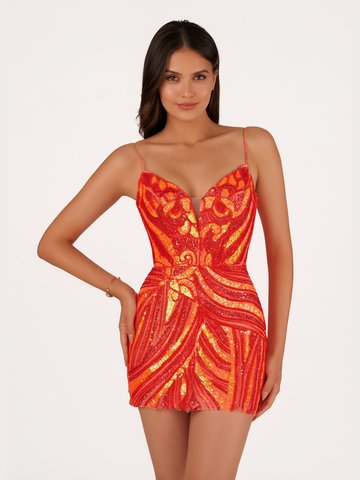} & \outimg{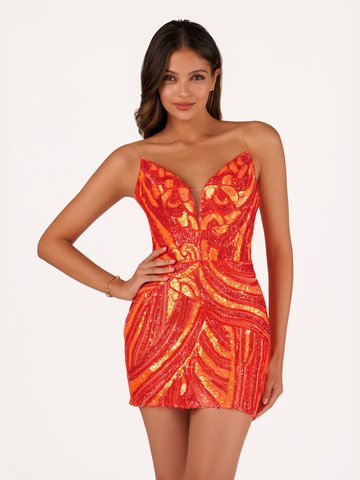} \\
\end{tabular}
\caption{Full outputs for the full-body target view ($s=1$) on two test gowns, seed 0 unless noted.
Left: trained without whole-garment targets, a model with scale conditioning only and SARA (seed 1) return a zoomed-in crop instead of the full view ($768{\times}1024$).
Right: at $1536{\times}2048$, the base model, the LoRA baseline and SARA all show the whole person; SARA uses the macro photo as an additional reference.
}
\label{fig:outputs}
\end{figure*}

\cref{tab:method} compares SARA with the LoRA baseline and the base model on the same garment region at the same size.
Both judges prefer SARA over the LoRA baseline on 18 of 40 items and the baseline on 6 ($p=0.023$); per gown, SARA is preferred on 12 and the baseline on 4.
Against the base model, SARA wins on 26 items and loses on 4 ($p=5.9{\times}10^{-5}$) and is preferred on 14 gowns against 1, a clearly larger margin than that of the LoRA baseline (23 against 10, $p=0.035$; 9 gowns against 1).
SARA keeps the design of the reference gown on 38 of 40 items, as often as the LoRA baseline, so its gain does not come from drifting to an easier design, and the local crops of the two models come from regions of similar size (125 and 117 pixels).
\cref{fig:zoomcrops} shows local crops and \cref{fig:outputs} full outputs.

\subsection{Analysis}
\label{sec:ablation}
\label{sec:collapse}
\paragraph{Whole-garment targets.}
Trained without the whole-garment targets, SARA learns to return a zoomed-in crop of the gown when asked for the full view, and so does a model with view conditioning and scale tags alone (\cref{fig:outputs}, left).
On the validation gowns, the garment box stays clear of the bottom image edge on 7 to 15 of 20 gowns across checkpoints from 500 to 8000 steps without whole-garment targets, and on 2 to 13 with scale conditioning alone, against 20 for the base model and 18 to 20 for the LoRA baseline, so no checkpoint of either is eligible.
With whole-garment targets, SARA keeps the full view on 20 of 20 validation gowns at step 2000 and 18 at step 4000, and on all 20 test gowns for both seeds.
\paragraph{Generation resolution.}
\label{sec:resolution}
The base model at $1536{\times}2048$ beats itself at $768{\times}1024$ on 38 of 40 items ($p=1.5{\times}10^{-9}$) with unchanged framing, because the crop then has about twice the side length (88 against 46 pixels).
At $768{\times}1024$, fine-tuned models frame the person closer than the base model, which gives their local crops more pixels and favors them in a pairwise comparison, so we compare all models at $1536{\times}2048$, where framing is similar.

\section{Limitations and Conclusion}
\label{sec:conclusion}

\paragraph{Limitations.}
\label{sec:limitations}
Local fidelity in this work is judged by two vision-language models from different families, each gated on control crops.
Judges of this kind can disagree on fine fabric structure, and their chance-corrected agreement on generated crops in the editor study was low ($\kappa=0.161$), so we count only labels on which both agree.
A controlled human study with trained raters requires institutional review board approval, which was pending at submission time; we will add it once approved.
\dataset comes from a single source, the test set has 20 usable gowns, and each model was trained with one seed, and automatic feature similarity does not separate SARA from the LoRA baseline at local scale, which is why we rely on judged comparisons.
The closed editors were run through their public interfaces at their largest settings, and their internal resolution is unknown.

\paragraph{Conclusion.}
Fine gown detail is the product, and it is what current editors lose when their outputs are viewed up close.
We proposed scale-aware reference attention, which tells the generator the zoom level of its output and of each reference, together with \dataset and zoom-crop evaluation.
Current editors, including the strongest commercial one we tested, lose or redraw local detail on most test gowns, and SARA is preferred over LoRA fine-tuning and over the base model at local scale while keeping the full view and the design of the gown.

\section*{LLM Usage Statement}
\label{sec:llm_usage}
Large language models were used in this work in three roles.
(i) Evaluation: two vision-language models, Claude Opus 5.5 and GPT-6.1, serve as judges of local fabric fidelity (\cref{sec:judged,sec:experiments}); a vision-language model (GPT-6.1) also labeled the views of the mannequin photos and checked the cross-domain pairs of \dataset (\cref{sec:dataset}).
Their prompts, controls and agreement are reported in the paper, and their outputs are treated as measurements with the limitations stated in \cref{sec:limitations}.
(ii) Research assistance: LLM-based coding agents (Claude and GPT models) wrote and ran most of the experiment code, data indexing scripts, statistical analyses and plotting code for the figures under the direction of the authors, who specified the experiments, decision rules and checks.
Analyses were recomputed by independent scripts, and an agent (GPT-6.1) checked every reference against its source.
(iii) Writing: an LLM-based agent drafted sections of the paper from the result files.
The authors reviewed the text, and every number in the paper is traced to a result file.
LLMs are not authors of this work, and the authors take full responsibility for its content.

\section*{Acknowledgments}
We thank the Phi Lab Foundation for providing computing resources and the experimental environment for this work.
{
    \small
    \bibliographystyle{ieeenat_fullname}
    \bibliography{refs_display}
}

\end{document}